\def\year{2020}\relax
\documentclass[letterpaper]{article} 
\usepackage{aaai20}  
\usepackage{times}  
\usepackage{helvet} 
\usepackage{courier}  
\usepackage[hyphens]{url}  
\usepackage{graphicx} 
\usepackage[ruled, vlined, linesnumbered]{algorithm2e} 
\usepackage{graphicx}  
\usepackage[normalem]{ulem}
\usepackage{amssymb}
\usepackage{dsfont}
\usepackage{multirow}

\usepackage{amssymb}
\usepackage{amsmath}
\newcommand{\Paragraph}[1]{\noindent\textbf{#1}}

\title{Probing the Natural Language Inference Task with Automated Reasoning Tools}
\author{Zaid Marji, Animesh Nighojkar, John Licato\\
Advancing Machine and Human Reasoning (AMHR) Lab\\
Department of Computer Science and Engineering\\
University of South Florida
}
\begin{document}

\maketitle

\begin{abstract}
The Natural Language Inference (NLI) task is an important task in modern NLP, as it asks a broad question to which many other tasks may be reducible: Given a pair of sentences, does the first entail the second? Although the state-of-the-art on current benchmark datasets for NLI are deep learning-based, it is worthwhile to use other techniques to examine the logical structure of the NLI task. We do so by testing how well a machine-oriented controlled natural language (Attempto Controlled English) can be used to parse NLI sentences, and how well automated theorem provers can reason over the resulting formulae. To improve performance, we develop a set of syntactic and semantic transformation rules. We report their performance, and discuss implications for NLI and logic-based NLP.
\end{abstract}


\section{Introduction}
\label{sec:intro}

Natural Language Inference (NLI) is the task of characterizing semantic relationships between sentences. Given a natural language sentence, called a \textit{premise}, what sort of relationship does it have with another sentence (called the \textit{hypothesis})? For example, consider the premise: \textit{``Two dogs are running through a field.''}. 
\begin{itemize}
    \item \textit{Entailment}: Given the premise, the hypothesis is certainly true \cite{Bowman2015}. \textit{``There are animals outdoors."}
    \item \textit{Contradiction}: Given the premise, the hypothesis is certainly false \cite{Bowman2015}. \textit{``The dogs are sitting on the couch."}
    \item \textit{Neutral}: Identification of any of the above two relationships requires more information, and therefore given the premise, the hypothesis may or may not be true \cite{Bowman2015}. \textit{``Some puppies are running to catch a stick."}
\end{itemize}

\noindent Current state-of-the-art results on datasets for the NLI task (e.g., \cite{Zhang2018IComprehension,Zhang2019Semantics-awareUnderstanding,Liu2019Multi-TaskUnderstanding}) rely almost exclusively on deep neural networks. As a reasoning task, it would seem that a logic-based approach could be useful---even if only to provide insights about the nature of the NLI task. In this paper, we implement and report our results using transformation rules to convert natural language into formal expressions that can be fed into automated theorem provers.


\subsection{SNLI}

The Stanford Natural Language Inference (SNLI) \cite{Bowman2015} dataset is a collection of labeled sentence pairs designed for the NLI task. It features 570,152 sentence pairs, which is twice as large as other datasets for the NLI task. It is entirely written by humans in a grounded and naturalistic context which enabled it to achieve high inter-annotator agreement: 98\% of the sentences selected for validation had at least 3 out of 5 independent raters agreeing on their classification. Unlike others, the SNLI dataset does not contain any sentences that are automatically generated or annotated.

Since all premises in the SNLI dataset were written by Amazon Mechanical Turk workers to describe a scenario in a picture, sentences tend to be more grammatical, making it ideal for our present task. In comparison, alternatives such as MultiNLI \cite{Williams2017} have a large number of ungrammatical or loosely structured sentences extracted from conversational or informal sources.

\subsection{ACE and APE}
Attempto Controlled English (ACE) is an English-based Controlled Natural Language (CNL), originally used for software specifications, whose focus eventually shifted to knowledge representation \cite{kuhn_survey_2014}. We chose ACE over other CNLs primarily for the following reasons:
\begin{enumerate}
    \item Unlike many other CNLs, ACE is not domain-specific.
    \item The syntax is purposely loosely-defined, thus providing more expressiveness \cite{Kuhn2010ControlledRepresentation}.
    \item All valid ACE sentences can be translated directly to First-Order Logic (FOL) \cite{Fuchs2008,fuchs2006,kuhn_survey_2014}. The ACE parser can output valid TPTP formulae directly, which we use in our prover.
    \item A strength of ACE is the abundance of related tools created by the Attempto\footnote{\url{http://attempto.ifi.uzh.ch}} group. Among these is the Attempto Parsing Engine (APE), a free tool to parse ACE and generate Discourse Representation Structures (DRS) \cite{Fuchs2011Discourse6.6}, TPTP \cite{Sutcliffe2009TheV3.5.0}, First-Order Logic (FOL), and parse trees.
\end{enumerate}

\section{Experiment Setup}

In order to study how well ACE can apply to the SNLI task, we set up a simple experiment.\footnote{Code: \url{https://github.com/AMHRLab/NLIwithACE/}} First, given a premise and hypothesis, we use APE to convert them into TPTP formulae. If this fails for either sentence, we apply a set of syntactic transformation rules (described below), and attempt the APE parse again. If it fails again, then we move on to the next pair, recording the percentage of premise or hypothesis sentences that successfully parse as the ``coverage level.''

Next, given TPTP formulae $P,H$ corresponding to the premise and hypothesis, we feed them into a first-order resolution-based prover to determine whether $H$ or its negation follow from $P$. If the prover outputs an affirmative answer to either of these (to keep run-times manageable, we automatically stop processing when 1500 clauses are created), then we give this as our classification. If not, then instead of outputting a classification of `neutral,' we apply a set of semantic transformation rules which produce a set of additional first-order formulas $\mathbf{A}$, capturing semantic information that may be of use to the prover. If $\mathbf{A} \cup \{P, \neg H\}$ resolves, output `entailment'; if $\mathbf{A} \cup \{P,H\}$ resolves, output `contradiction'; Finally, we generate a set of semantic rules based on the \textit{closed-world assumption} \cite{Mueller2015}, yielding another set of first-order formulae $\mathbf{N}$. If the result is still inconclusive, output is `neutral'. Pseudocode for this algorithm is shown in Algorithm \ref{alg:pseudocode}.


\subsection{Syntactic Rewrite Rules}

Abdelaal \cite{Abdelaal2019} proposed leveraging controlled natural languages (CNLs) to improve knowledge extraction. There are two major types of CNLs: human-oriented and machine-oriented. Human-oriented CNLs are usually developed as style-guides with the aim of avoiding complex grammatical structure and reducing potential ambiguity. On the other hand, machine-oriented CNLs aim to provide a user-friendly machine-processable format. They tend to be much more restricted compared to human-oriented CNLs. ACE is a machine-oriented CNL, while Basic English (the CNL that is adopted by Simple English Wikipedia (SEW)) is a human-oriented CNL. Abdelaal's thesis explores the possibility of extracting knowledge from SEW articles by converting the human-oriented CNL of SEW into a machine-oriented CNL, namely ACE. The proposed solution is to rewrite sentences by following rules such that they become valid ACE sentences. The thesis proposed 10 rewriting rules, along with psuedocode implementations.

In this paper, our goal is to leverage automated reasoning tools for the NLI task. Our hypothesis is that we can convert SNLI sentences to ACE using rewrite rules, and use FOL theorem provers to detect entailment and contradiction relationships in a non-trivial portion of the SNLI dataset. Therefore, we used the rules described in \cite{Abdelaal2019} as a starting point, and implemented our own versions in Python. These rules, which we call \textit{syntactic rewrite rules}, are meant only to change the syntax of the sentences (and not their semantics, whenever possible) so that they can be ACE. However, after initial experimentation, we found that the specific needs of SNLI required significant changes. High-level descriptions of the final versions of all syntactic rewrite rules we implemented are as follows:

\Paragraph{R1: Noun/adjective phrases.} For each sentence pair, the SNLI dataset provides part-of-speech tags and constituency parses, which we use directly. If there is a NP consisting of a sequence of JJs followed by a NN or NNS, then we attach POS markers `a:' to each JJ and `n:' to the noun. POS markers are used by APE to identify words that may not be in its vocabulary. If there are multiple JJs, we make them an adjective phrase using conjunctions, so that \texttt{(NP [(DT d)] (JJ adj1) (JJ adj2) ... (JJ adjn) (NN[S] n))} transforms into \texttt{(NP [(DT d)] (ADJP (JJ adj1) (CC and) (JJ adj2) (CC and) ... (JJ adjn)) (NN[S] n))}. In contrast, \cite{Abdelaal2019} used hyphens to conjoin adjective phrases, making it difficult to reason about individual adjectives.


\Paragraph{R2: Co-reference resolution.} ACE does not allow pronouns. We use Stanford's CoreNLP server to identify coreference chains within each sentence. If the chain contains a proper noun, that noun is considered the chain's name; otherwise, a default and unique name is used. All words in that chain are replaced with the chain's name. The POS prefix ``p:'' is then added to all names.

However, this will lose some information. For example, the sentence ``John loves his wife and she is laughing at him.'' will be replaced with ``p:John loves p:John's wife and p:DefaultName0 is laughing at p:John.'' But we no longer will know that `p:DefaultName0' is John's wife. Thus, all words in the chain marked as singular NOMINAL are converted into additional sentences. In the above example, the sentence ``p:DefaultName0 is p:John's wife.'' is appended.

\Paragraph{R3: Past Tense Verbs:} Since ACE does not parse past tense, replace past tense verbs (VBD/VBN) with present tense, using pattern.en.\footnote{\url{https://www.clips.uantwerpen.be/pages/pattern}}

\Paragraph{R4: Cardinals and Ordinals.} Any occurrences of numbers 1, ..., 10 are replaced with the words `one,' ..., `ten.' Ordinals `1st,' ..., `10th' are replaced with `first,' ..., `tenth.'

\Paragraph{R5: Predeterminers.} All predeterminers (as identified by SNLI-provided POS tags) are removed.

\Paragraph{R6: Adverb phrase ordering.} If a VP has an ADVP preceding a verb, swap their order. 

\Paragraph{R7: Adverb conjunctions.} If an ADVP has multiple adverbs joined by 'but' or 'yet', join them instead with 'and'. 

\Paragraph{R8: Removing Present Continuous.} Present and past continuous form (``is/are/was/were \texttt{Verb}-ing") are replaced with simple present tense. This helps to simplify formulae; e.g., APE transforms ``Nobody is working" into a TPTP formula with a superfluous equality subformula, whereas ``Nobody works" does not.


\subsection{Semantic Rules}
After applying the syntactic rules, we apply a small set of \textit{semantic rules}, which are meant to enhance the sentences with semantic information in a way that preserves the entailment relationship, and otherwise aids the inference step, but does not necessarily preserve the meaning of the individual sentences. They also differ from syntactic rewrite rules in that they consider both the premise and hypothesis, rather than each separately. They are not expected to affect coverage.


\Paragraph{S1: Noun Hypernyms.} Knowing that certain types of nouns are hypernyms of others is necessary for the SNLI task, but rewriting the premise and hypothesis to include this information can be highly inefficient. For example, consider the sentence pair ``A woman hugs a boy''/``A person hugs a boy''. It would not be immediately clear how to rewrite the sentences to ensure that the relationship between the two is preserved. We might detect that `person' is a hypernym of both `woman' and `boy,' and this might cause us to overshoot and replace the premise with ``A person hugs a person." This would cause the entailment between the premise and conclusion to no longer hold.


Instead, we keep the premise and hypothesis the same, and use hypernym information to construct additional first-order formulae that are included in the set \textbf{A} that is used in the first-order resolution step. Given a sentence pair, we scan the premise and conclusion for all nouns, and save their singular forms $S$. For any two $n_1, n_2 \in S$, we use WordNet \cite{Miller1990IntroductionDatabase} to check if $n_2$ is a hypernym of $n_1$. For every such hypernym pair found, we add the following formula to \textbf{A}:
\begin{equation}
   \forall_x n_1(x) \rightarrow n_2(x) 
\end{equation}
When APE parses singular nouns, it will typically introduce a predicate corresponding to the noun itself, which would align with the formula above. For any two nouns such that we can not show that one is a hypernym of the other, we add the following formula to the set \textbf{N}:
\begin{equation}
   \forall_x n_1(x) \leftrightarrow \neg n_2(x) 
\end{equation}


\Paragraph{S2: Verb hypernyms.} We want to be able to capture the entailment in sentence pairs such as ``A young boy sprints by the beach''/``A boy runs.'' Like with \textbf{S1}, we collect all pairs of verbs $v_1,v_2$ and compare them in their infinitive forms, as determined by pattern.en. Because we are not aware whether the verb is transitive or not, we add multiple formulas into \textbf{A} for each verb where $v_2$ is a hypernym of $v_1$:
\begin{equation} \label{eq:HB2}
  \begin{aligned}
    \forall_{a,b,c} \; predicate2(&a, v_1, b, c) \rightarrow \\
        &predicate2(a, v_2, b,c) \\
  \end{aligned}
\end{equation}
Here, \textit{predicate2} is the formula APE uses to encode transitive verbs with one object; we also create analogous formulae for \textit{predicate1} (intransitive verbs). Likewise, we create analogous formulae in \textbf{N} for all verb pairs $v_1,v_2$ when we determine neither is a hypernym of the other according to WordNet:
\begin{equation} \label{eq:HB2}
  \begin{aligned}
    \forall_{a,b,c} \; &predicate2(a, v_1, b, c) \leftrightarrow \\
        &\neg predicate2(a, v_2, b,c) \\
  \end{aligned}
\end{equation}

\begin{algorithm}[t]
\SetAlgoLined
Load the SNLI sentence pair $(P, H)$\;
$(P_{t}, H_{t}) =$ attempted APE conversion of $(P, H)$ into TPTP\;
\uIf{$P$ or $H$ fails to parse}{Apply syntactic rewrite rules and try again\;
 \uIf{one of them fails to parse}{Abandon and go to next sentence pair}}
Feed $(P_{t}, H_{t})$ into prover\;
\uIf{prover guessed `entailment' or `contradiction'}{Assess guess and go to next sentence pair}
\Else{Apply semantic rules to $(P_{t}, H_{t})$\;
 Store additional formulae produced as a result of those rules in $\mathbf{A}$ and $\mathbf{N}$\;
 Feed $(P_{t}, H_{t}, \mathbf{A})$ into prover\;
  \uIf{prover guessed 'entailment' or 'contradiction'}{Assess guess and go to next sentence pair}
  \Else{Feed $(P_{t}, H_{t}, \mathbf{A}, \mathbf{N})$ into prover\;
  Assess guess and go to next sentence pair}}
 \caption{Algorithm Used to Process SNLI}
 \label{alg:pseudocode}
\end{algorithm}










\section{Results and Analysis}

To begin our evaluation, we first determined the coverage level and classification accuracies for the SNLI development set (10,000 sentence pairs) without any rewrite rules applied. After applying the syntactic rewrite rules, coverage increased dramatically, going from 7.06\% (before rule application) to 16.61\%. 

The classification accuracies, considering only sentence pairs which APE was able to parse after syntactic rules but without semantic rules, are presented in Table \ref{tbl:A1}. As expected, the overall classification accuracy alone (roughly 28.7\%) is not comparable to current state-of-the-art systems, and is not above random baseline. However, it is worthwhile to note that the top-left cell in Table \ref{tbl:A1} is 100\%; this means that when a pair of SNLI sentences are ACE, if an automated theorem prover guessed that the pair was an entailment, it was correct every single time. 

We expect similar results for predictions of contradiction; however, prior to the application of semantic rules, none were made. This is due to the fact that the ACE parses alone of the largely descriptive sentences extremely rarely produced formulae containing negations. In other words, SNLI sentences are overwhelmingly descriptions of what \textit{is} true, and almost never about what \textit{is not} true. It was our hope this limitation would be addressed by the semantic rules. However, the results (Table \ref{tbl:A2}) are disappointing: Contradiction predictions are indeed more common, but the accuracy (37.5\%) is barely better than random baseline.


\begin{table}[t]
\centering
\begin{tabular}{cc|c|c|c|}
\cline{3-5}
\textbf{} & \textbf{} & \multicolumn{3}{c|}{\textbf{Predicted}} \\ \cline{3-5} 
\textbf{} &  & \textbf{E} & \textbf{N} & \textbf{C} \\ \hline
\multicolumn{1}{|c|}{\multirow{3}{*}{\rotatebox{90}{\textbf{Actual}}}} & \textbf{E} & 1.0 & 0.36 & 0.0 \\ \cline{2-5} 
\multicolumn{1}{|c|}{} & \textbf{N} & 0.0 & 0.256 & 0.0 \\ \cline{2-5} 
\multicolumn{1}{|c|}{} & \textbf{C} & 0.0 & 0.384 & 0.0 \\ \hline
\end{tabular}
\caption{Confusion matrix without semantic transformations}
\label{tbl:A1}
\end{table}

\begin{table}[t]
\centering
\begin{tabular}{cc|c|c|c|}
\cline{3-5}
\textbf{} & \textbf{} & \multicolumn{3}{c|}{\textbf{Predicted}} \\ \cline{3-5} 
\textbf{} &  & \textbf{E} & \textbf{N} & \textbf{C} \\ \hline
\multicolumn{1}{|c|}{\multirow{3}{*}{\rotatebox{90}{\textbf{Actual}}}} & \textbf{E} & 0.383 & 0.439 & 0.406 \\ \cline{2-5} 
\multicolumn{1}{|c|}{} & \textbf{N} & 0.228 & 0.341 & 0.219 \\ \cline{2-5} 
\multicolumn{1}{|c|}{} & \textbf{C} & 0.389 & 0.22 & 0.375 \\ \hline
\end{tabular}
\caption{Confusion matrix with semantic transformations}
\label{tbl:A2}
\end{table}

In the process of reviewing the output of our algorithm, we found that some sentence pairs in the SNLI dataset were mislabeled. For example, the sentence pair \textit{``Two young girls hug.''} and \textit{``The girls are happy.''} is classified as \textit{entailment} in the dataset, but our algorithm guessed \textit{neutral}, due to not making the assumption that individuals who hug are happy. Whether such assumptions are warranted is an interesting question, but it is interesting to think that such assumptions can be brought to light by logic-based approaches.

\section{Conclusions and Future Work}

This short paper presents work-in-progress, and represents exciting research possibilities. Although the approach used here does not obtain higher accuracy on the NLI task than current exclusively deep learning-based approaches, this preliminary work showed how the coverage of SNLI sentences can be dramatically improved with simple, semantics-preserving, syntactic rewrite rules.

Further research into how to improve the rewriting rules described here can offer interesting insights into future developments of CNLs, automated reasoning, and the NLI task itself (e.g., the simple logic-based approach in this paper showed that a non-insignificant number of mislabeled examples exist in the SNLI dataset). One potential future direction would be to compare the approach presented in this paper to other logic-based approaches, such as natural logic (e.g. LangPro \cite{abzianidze2017langpro}, Monalog \cite{hu2019monalog}, or NaturalLI \cite{angeli2014naturalli}), and explore the potential of combining different approaches into a consolidated solution. We hope to use the present work as a baseline against which future work can compare.


\section{Acknowledgements}

\textit{This material is based upon work supported by the Air Force Office of Scientific Research under award numbers FA9550-17-1-0191 and FA9550-18-1-0052. Any opinions, findings, and conclusions or recommendations expressed in this material are those of the authors and do not necessarily reflect the views of the United States Air Force.}


\bibliographystyle{aaai}
\bibliography{john, animesh, extra}

\begin{thebibliography}{}

\bibitem[\protect\citeauthoryear{Abdelaal}{2019}]{Abdelaal2019}
Abdelaal, H.~S.
\newblock 2019.
\newblock {\em Knowledge Extraction from Simplified Natural Language Text}.
\newblock Ph.D. Dissertation, National University of Ireland, Galway.

\bibitem[\protect\citeauthoryear{Abzianidze}{2017}]{abzianidze2017langpro}
Abzianidze, L.
\newblock 2017.
\newblock {L}ang{P}ro: Natural language theorem prover.
\newblock In {\em Proceedings of the 2017 Conference on Empirical Methods in
  Natural Language Processing: System Demonstrations},  115--120.
\newblock Copenhagen, Denmark: Association for Computational Linguistics.

\bibitem[\protect\citeauthoryear{Angeli and
  Manning}{2014}]{angeli2014naturalli}
Angeli, G., and Manning, C.~D.
\newblock 2014.
\newblock Naturalli: Natural logic inference for common sense reasoning.
\newblock In {\em Proceedings of the 2014 conference on empirical methods in
  natural language processing (EMNLP)},  534--545.

\bibitem[\protect\citeauthoryear{Bowman \bgroup et al\mbox.\egroup
  }{2015}]{Bowman2015}
Bowman, S.~R.; Angeli, G.; Potts, C.; and Manning, C.~D.
\newblock 2015.
\newblock A large annotated corpus for learning natural language inference.
\newblock In {\em Proceedings of the 2015 Conference on Empirical Methods in
  Natural Language Processing (EMNLP)}.
\newblock Association for Computational Linguistics.

\bibitem[\protect\citeauthoryear{Fuchs, Kaljurand, and Kuhn}{2008}]{Fuchs2008}
Fuchs, N.~E.; Kaljurand, K.; and Kuhn, T.
\newblock 2008.
\newblock {Attempto Controlled English for Knowledge Representation}.
\newblock In Baroglio, C.; Bonatti, P.~A.; Ma{\l}uszy{\'{n}}ski, J.; Marchiori,
  M.; Polleres, A.; and Schaffert, S., eds., {\em Reasoning Web: 4th
  International Summer School 2008, Venice, Italy, September 7-11, 2008,
  Tutorial Lectures}. Berlin, Heidelberg: Springer Berlin Heidelberg.
\newblock  104--124.

\bibitem[\protect\citeauthoryear{Fuchs, Kaljurand, and
  Kuhn}{2011}]{Fuchs2011Discourse6.6}
Fuchs, N.~E.; Kaljurand, K.; and Kuhn, T.
\newblock 2011.
\newblock {Discourse Representation Structures for ACE 6.6}.
\newblock Technical Report August.

\bibitem[\protect\citeauthoryear{Fuchs, Kaljurand, and
  Schneider}{2006}]{fuchs2006}
Fuchs, N.~E.; Kaljurand, K.; and Schneider, G.
\newblock 2006.
\newblock Attempto controlled english meets the challenges of knowledge
  representation, reasoning, interoperability and user interfaces.
\newblock In {\em FLAIRS Conference}, volume~12,  664--669.

\bibitem[\protect\citeauthoryear{Hu \bgroup et al\mbox.\egroup
  }{2019}]{hu2019monalog}
Hu, H.; Chen, Q.; Richardson, K.; Mukherjee, A.; Moss, L.~S.; and Kuebler, S.
\newblock 2019.
\newblock Monalog: a lightweight system for natural language inference based on
  monotonicity.
\newblock {\em arXiv preprint arXiv:1910.08772}.

\bibitem[\protect\citeauthoryear{Kuhn and
  Hess}{2010}]{Kuhn2010ControlledRepresentation}
Kuhn, T., and Hess, P. D.~M.
\newblock 2010.
\newblock {Controlled English for Knowledge Representation}.
\newblock {\em Faculty of Economics, Business Administration and Information
  Technology} Doctor(November):246.

\bibitem[\protect\citeauthoryear{Kuhn}{2014}]{kuhn_survey_2014}
Kuhn, T.
\newblock 2014.
\newblock {A Survey and Classification of Controlled Natural Languages}.
\newblock {\em Computational Linguistics} 40(1):121--170.

\bibitem[\protect\citeauthoryear{Liu \bgroup et al\mbox.\egroup
  }{2019}]{Liu2019Multi-TaskUnderstanding}
Liu, X.; He, P.; Chen, W.; and Gao, J.
\newblock 2019.
\newblock {Multi-Task Deep Neural Networks for Natural Language Understanding}.
\newblock  4487--4496.

\bibitem[\protect\citeauthoryear{Miller \bgroup et al\mbox.\egroup
  }{1990}]{Miller1990IntroductionDatabase}
Miller, G.~A.; Beckwith, R.; Fellbaum, C.; Gross, D.; and Miller, K.~J.
\newblock 1990.
\newblock {Introduction to wordnet: An on-line lexical database}.
\newblock {\em International Journal of Lexicography} 3(4):235--244.

\bibitem[\protect\citeauthoryear{Mueller}{2015}]{Mueller2015}
Mueller, E.~T.
\newblock 2015.
\newblock {\em {Commonsense Reasoning: An Event Calculus Based Approach}}.
\newblock Morgan Kaufmann, second edition edition.

\bibitem[\protect\citeauthoryear{Sutcliffe}{2009}]{Sutcliffe2009TheV3.5.0}
Sutcliffe, G.
\newblock 2009.
\newblock {The TPTP problem library and associated infrastructure : ttthe FOF
  and CNF Parts, v3.5.0}.
\newblock {\em Journal of Automated Reasoning} 43(4):337--362.

\bibitem[\protect\citeauthoryear{Williams, Nangia, and
  Bowman}{2017}]{Williams2017}
Williams, A.; Nangia, N.; and Bowman, S.~R.
\newblock 2017.
\newblock A broad-coverage challenge corpus for sentence understanding through
  inference.
\newblock {\em CoRR} abs/1704.05426.

\bibitem[\protect\citeauthoryear{Zhang \bgroup et al\mbox.\egroup
  }{2018}]{Zhang2018IComprehension}
Zhang, Z.; Wu, Y.; Li, Z.; He, S.; and Zhao, H.
\newblock 2018.
\newblock {I Know What You Want: Semantic Learning for Text Comprehension}.

\bibitem[\protect\citeauthoryear{Zhang \bgroup et al\mbox.\egroup
  }{2019}]{Zhang2019Semantics-awareUnderstanding}
Zhang, Z.; Wu, Y.; Zhao, H.; Li, Z.; Zhang, S.; Zhou, X.; and Zhou, X.
\newblock 2019.
\newblock {Semantics-aware BERT for Language Understanding}.
\newblock (2017).

\end{thebibliography}

\end{document}